\documentclass[runningheads]{llncs}

\usepackage{eccv}

\usepackage{eccvabbrv}
\usepackage{wrapfig} 
\usepackage{graphicx}
\usepackage{booktabs}
\usepackage{multirow}
\usepackage{algorithm}
\usepackage{algorithmic}
\usepackage{enumitem}

\usepackage[accsupp]{axessibility}  

\usepackage{hyperref}

\usepackage{orcidlink}

\begin{document}

\title{V-tableR1: Process-Supervised Multimodal Table Reasoning with Critic-Guided Policy Optimization} 

\titlerunning{V-tableR1: Process-Supervised Multimodal Table Reasoning}



\author{
Yubo Jiang$^{1,2}$ \quad
Yitong An$^{1}$ \quad
Xin Yang$^{2}$ \quad
Abudukelimu Wuerkaixi$^{2}$ \quad
Xuxin Cheng$^{2}$ \quad
Fengying Xie$^{1,3}$ \\
Zhiguo Jiang$^{3}$ \quad
Cao Liu$^{2}$ \quad
Ke Zeng$^{2~\dagger}$ \quad
Haopeng Zhang$^{1,3~\dagger}$\\[0.5em]
$^{1}$School of Astronautics, Beihang University, Beijing 102206, China\\
$^{2}$Longcat Interaction Team, Meituan, Beijing 100102, China \\
$^{3}$Tianmushan Laboratory, Beihang University, Hangzhou 311115, China\\[0.5em]
{\tt\small \{jbond0409, zhanghaopeng\}@buaa.edu.cn (Y.J., H.Z.)}
}
\maketitle

\begin{abstract}
  We introduce V-tableR1, a process-supervised reinforcement learning framework that elicits rigorous, verifiable reasoning from multimodal large language models (MLLMs). Current MLLMs trained solely on final outcomes often treat visual reasoning as a black box, relying on superficial pattern matching rather than performing rigorous multi-step inference. While Reinforcement Learning with Verifiable Rewards could enforce transparent reasoning trajectories, extending it to visual domains remains severely hindered by the ambiguity of grounding abstract logic into continuous pixel space. We solve this by leveraging the deterministic grid structure of tables as an ideal visual testbed. V-tableR1 employs a specialized critic VLM to provide dense, step-level feedback on the explicit visual chain-of-thought generated by a policy VLM. To optimize this system, we propose Process-Guided Direct Alignment policy Optimization (PGPO), a novel RL algorithm integrating process rewards, decoupled policy constraints, and length-aware dynamic sampling. Extensive evaluations demonstrate that V-tableR1 explicitly penalizes visual hallucinations and shortcut guessing. By fundamentally shifting multimodal inference from black-box pattern matching to verifiable logical derivation, V-tableR1 4B establishes state-of-the-art accuracy among open-source models on complex tabular benchmarks, outperforming models up to 18$\times$ its size and improving over its SFT baseline by up to 10.6\% absolute accuracy.
\end{abstract}

\section{Introduction}
\label{sec:intro}

Vision-language models (VLMs) fail systematically on tabular reasoning.
Given a table image and a question requiring precise cell localization or multi-hop aggregation, state-of-the-art VLMs frequently produce confident yet incorrect answers\cite{favero2024multi,rawte2025defining,zhao2025cot,zhang2024vision,zhou2022learning,laurenccon2024matters,zhu2023minigpt}.
The failure is not perceptual---these models can read individual cell values correctly\cite{su2025crossing,zhang2023multimodal}.
Rather, the failure is \emph{inferential}: models shortcut multi-step reasoning by exploiting statistical surface patterns, bypassing the precise cell-lookup and logical chaining that a correct answer demands.
We trace this failure to a shared limitation of dominant training paradigms: they supervise only the final answer, leaving the intermediate reasoning trajectory entirely unconstrained\cite{robinson2021can,wang2024chain,geirhos2020shortcut,wu2025table,sun2020tableqa}.

\begin{figure}[t]
  \centering
  \includegraphics[width=\linewidth]{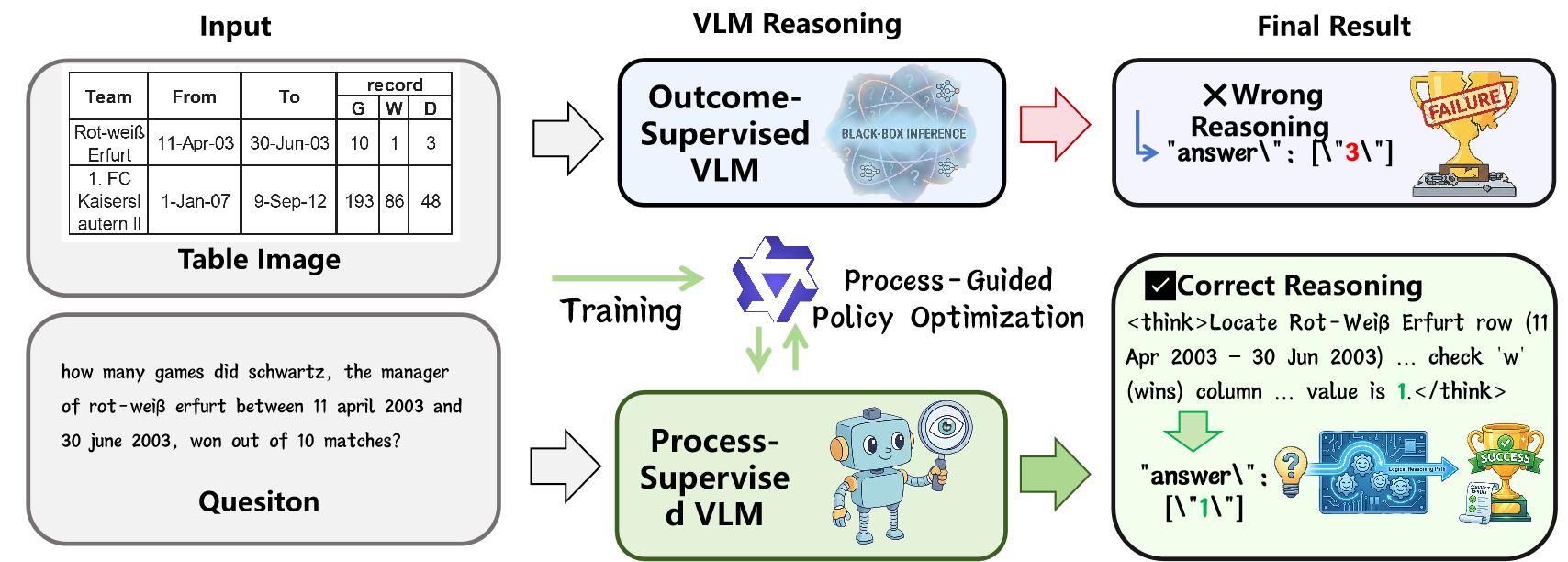}
  \caption{Comparison between outcome-supervised and process-supervised VLMs for table reasoning.
  Process supervision enables the model to accurately locate relevant cells and follow logical steps,
  leading to correct answers where black-box inference fails.}
  \label{fig:process_supervision}
\end{figure}
Even when augmented with explicit Chain-of-Thought trajectories, Supervised Fine-Tuning (SFT) relies on teacher forcing, which evaluates reasoning as a static sequence of tokens rather than a verifiable cognitive process\cite{zhang2025survey}. Under this paradigm, the model learns the syntax of reasoning but struggles to rigorously ground each step to visual evidence. Crucially, SFT lacks an active mechanism to penalize spurious logic; a model that hallucinates intermediate steps but luckily guesses the correct final answer receives no corrective signal during inference.

While Reinforcement Learning with Verifiable Rewards (RLVR) has achieved massive success in text-based math and coding, extending RLVR to multimodal tasks remains notoriously difficult\cite{su2025crossing,mroueh2025reinforcement}.
Natural image visual question answering (VQA) often involves subjective answers, making rigorous reward construction nearly impossible and highly susceptible to reward hacking\cite{zhang2024vision,zhou2022learning}.

Unlike natural scenes, tables encode strict, rule-bound logic within a rigid grid.
Both the intermediate reasoning trajectories (e.g., cell coordinate localization) and the final discrete answers are rigorously verifiable\cite{zhao2024tabpedia,antol2015vqa}.
This makes tabular reasoning an ideal, deterministic testbed for exploring multimodal RLVR.
Currently, the advancement of tabular vision models is heavily constrained by their reliance on SFT and RL methods driven by sparse outcome rewards. Both paradigms critically lack high-quality supervision for the intermediate reasoning process. Consequently, they retain a fundamental flaw: a reasoning trajectory that blindly stumbles to the correct answer via spurious correlations receives the exact same training signal as one that reasons correctly at every step \cite{geirhos2020shortcut,wu2025table,sun2020tableqa}. Figure~\ref{fig:process_supervision} makes this contrast concrete---outcome-supervised VLMs treat inference as a black box and fail on queries requiring explicit cell localization, while process-supervised models expose and verify each reasoning step, yielding correct answers \cite{wang2024rethinking}.

To overcome the inherent limitations of rigid SFT sequence modeling and outcome-driven reward hacking, we introduce \textbf{V-tableR1}, a process-supervised reinforcement learning framework designed to elicit rigorous, verifiable reasoning from MLLMs on tabular tasks.
V-tableR1 trains a VLM to produce an explicit, verifiable chain-of-thought that localizes cells, validates intermediate values, and resolves multi-hop dependencies before committing to a final answer.
Instead of relying on sparse outcome rewards that encourage shortcut learning, we propose \textbf{Process-Guided Direct Alignment Policy Optimization (PGPO)}. 
PGPO is a novel GRPO-based\cite{shao2024deepseekmath} reinforcement learning algorithm that seamlessly integrates decoupled policy constraints with dynamic, length-aware data sampling.
By directly injecting fine-grained, step-level feedback into the optimization objective, PGPO reinforces mathematically rigorous reasoning paths while explicitly penalizing hallucinatory leaps.

\medskip\noindent\textbf{Contributions.}
\begin{itemize}[itemsep=2pt,topsep=0pt,parsep=0pt]
  \item Visual Chain-of-Thought Dataset. We construct a large-scale dataset of table-question pairs with step-level reasoning annotations. These annotations explicitly link logical operations to visual anchors (e.g., cell coordinates and row/column identifiers), transforming implicit intuition into rigorous, verifiable training signals.
  \item Process Evaluation via critic VLM. We design a specialized critic VLM that functions as a continuous process verifier. Rather than treating rewards as black-box binary signals, the critic evaluates the logical fidelity of each generated step to produce a dense process score. This score dynamically gates the outcome reward: flawless reasoning receives a golden-case bonus, while unreliable chains that happen to guess the correct answer are conservatively penalized.
  \item Process-Guided Direct Alignment Policy Optimization (PGPO). We propose PGPO, a novel RL algorithm tailored for long-horizon visual reasoning. Building upon the stability of Direct Alignment Policy Optimization (DAPO) and the dynamic sampling efficiency of Length-aware Sampling for Policy Optimization (LSPO), PGPO leverages fine-grained process feedback to optimize the policy model. This unified objective fundamentally shifts the learning paradigm from guessing correct outcomes to mastering verifiable, robust reasoning trajectories.
\end{itemize}

\section{Related Work}
\subsection{Vision-Language Models for Structured Reasoning}
Recent advances in multimodal foundation models expand visual understanding from natural scenes to dense, structured documents like tables and charts.
Unlike natural images, tables demand precise spatial-semantic alignment to decode hierarchical headers and rigid grid layouts\cite{titiya2025mmtbench}.
Standard approaches rely heavily on Supervised Fine-Tuning (SFT), utilizing massive image-text datasets to align visual encoders with language models\cite{zheng2024multimodal,yang2022tableformer}.
While SFT improves superficial structural recognition, it operates under an outcome-supervised paradigm.
Models maximize the likelihood of the final token sequence, treating the intermediate logical derivation as a black box.
Consequently, when faced with complex queries requiring multi-hop aggregation or precise numerical comparison, these models rely on memorized lexical correlations rather than genuine comprehension, frequently hallucinating plausible but incorrect answers\cite{ji2024tree}.
We diverge from this paradigm by enforcing explicit Visual Chain-of-Thought reasoning, transforming table understanding from opaque pattern matching into a transparent, step-by-step verifiable process\cite{chen2024visual}.

\subsection{Reinforcement Learning with Verifiable Rewards}
Reinforcement Learning with Verifiable Rewards (RLVR) drives recent breakthrough reasoning capabilities in text-only language models, unlocking test-time scaling and extended chain-of-thought paradigms\cite{yue2025vapo}.
Algorithms like Group Relative Policy Optimization (GRPO) demonstrate that optimizing directly against rule-based, verifiable outcomes elicits sophisticated behaviors such as autonomous self-correction\cite{shao2024deepseekmath,vojnovic2025alignment}.
Subsequent innovations refine this foundation; notably, Direct Alignment Policy Optimization (DAPO) mitigates gradient variance and stabilizes long-context reasoning by decoupling clipping constraints\cite{yu2025dapo,yue2025vapo}.
However, translating these successes to the multimodal domain remains challenging.
In pure text, a mathematical proof is logically absolute; in vision, grounding a logical step to a continuous pixel space introduces ambiguity that resists strict reward verification and invites reward hacking\cite{chen2024visual}.
We identify table reasoning as the ideal multimodal testbed for RLVR because tables possess the deterministic structure necessary for rigorous verification.
Furthermore, relying solely on sparse outcome rewards in visual tasks exacerbates shortcut learning\cite{chen2025lspo}.
Our PGPO algorithm directly addresses this limitation by building upon the stable foundations of GRPO variants, injecting dense, process-level feedback into the optimization objective to guide the model away from spurious correlations.

\subsection{Process Supervision and Step-Level Feedback}
To mitigate the limitations of sparse outcome rewards, process-supervised learning assigns credit directly to intermediate reasoning steps.
In text domains like mathematical theorem proving and code generation, Process Reward Models (PRMs) successfully guide policies toward robust reasoning paths by penalizing early logical errors before they compound\cite{wang2025visualprm,cui2025process}.
Yet, applying process supervision to visual tasks proves remarkably difficult.
The primary bottleneck lies in grounding abstract logical steps to specific visual anchors, such as pixel coordinates or cell boundaries.
Previous multimodal efforts circumvent this alignment problem by relying on heuristic rule-checkers restricted to highly constrained synthetic environments\cite{zhang2025lessons,jiao2024learning}.
We bridge this critical gap by training a specialized critic VLM to serve as a continuous process verifier\cite{cui2025process}.
By dynamically gating the outcome reward based on the logical fidelity of the explicit visual reasoning chain, our framework provides the first scalable mechanism for dense process supervision in multimodal table reasoning\cite{lu2022learn}.

\section{Method}
\label{sec:method}

We introduce \textbf{V-tableR1}, a process-supervised reinforcement learning framework designed to elicit rigorous, verifiable reasoning from multimodal large language models (MLLMs) on tabular tasks.
Our approach addresses the critical vulnerability of outcome-only supervision in visual domains: the propensity for models to hallucinate intermediate steps while stumbling upon correct final answers.
Unlike natural images where objects lack definitive structural boundaries, tables impose a rigid, grid-based logic.
This intrinsic determinism makes table reasoning an ideal testbed for Reinforcement Learning with Verifiable Rewards (RLVR)---every intermediate cell reference and aggregation step can be rigorously verified\cite{wang2024chain,zhang2023multimodal}.

\begin{figure}[t]
  \centering
  \includegraphics[width=\linewidth]{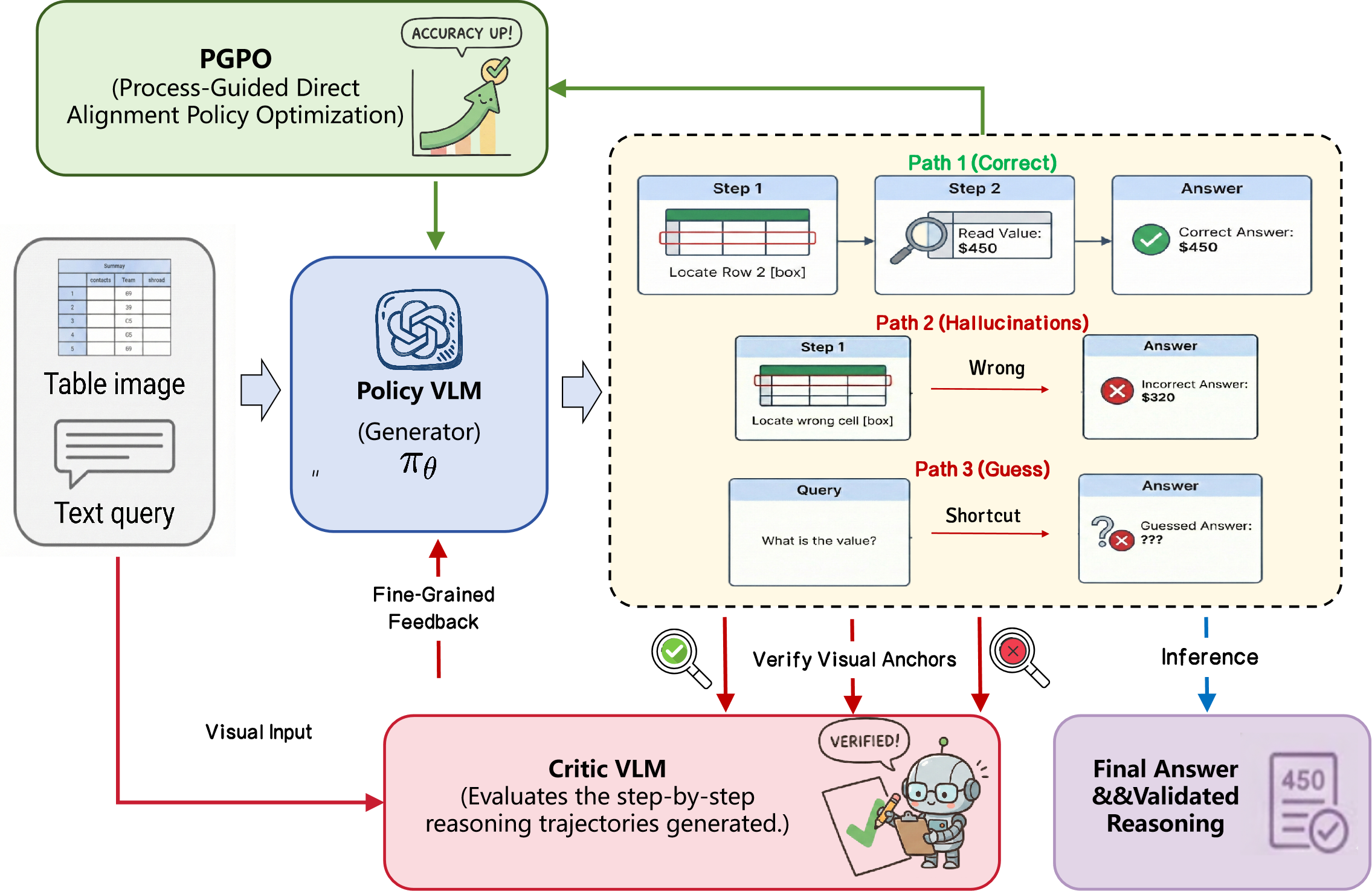}
  \caption{Overview of the V-tableR1 framework. The policy VLM generates an explicit Visual Chain-of-Thought (V-CoT) over the table image. The critic VLM verifies the visual anchors to distinguish between rigorous inference (Path 1), visual hallucination (Path 2), and shortcut guessing (Path 3). This dense process feedback is then integrated into the PGPO algorithm to optimize the policy.}
  \label{fig:framework}
\end{figure}

To operationalize this, V-tableR1 decomposes inference into two distinct interacting models (illustrated in Figure~\ref{fig:framework}): a policy VLM ($\pi_\theta$) that generates an explicit Visual Chain-of-Thought (V-CoT) and a critic VLM that evaluates the logical and visual fidelity of each intermediate step.
We optimize $\pi_\theta$ using a novel algorithm, Process-Guided Direct Alignment Policy Optimization (PGPO), which unifies the gradient stability of Direct Alignment Policy Optimization (DAPO) and the sampling efficiency of Length-aware Sampling for Policy Optimization (LSPO), while crucially injecting dense process-level feedback.

\paragraph{Visual Chain-of-Thought Generation}

Let an input query be a tuple $(x, q)$, where $x$ is the table image and $q$ is the natural language question.
The policy VLM generates a sequence of reasoning steps culminating in a final answer: $y = (s_1, a_1, \dots, s_K, a_K, \text{ans})$.
To break the black-box nature of tabular reasoning, we enforce the generation of Visual Anchors ($a_k$) alongside logical steps ($s_k$).
Rather than operating in raw pixel space, the policy VLM is supervised during an initial SFT phase to ground its reasoning in relative grid/logical coordinates (e.g., \texttt{<cell: Row 2, Col 3>}).
By explicitly outputting these coordinates and their corresponding cell values before executing any arithmetic or comparative logic, the model produces a V-CoT trajectory that is mathematically and structurally verifiable\cite{wang2025multimodal,zhang2025survey,zheng2024multimodal}.

As depicted in Figure~\ref{fig:framework}, this explicit generation exposes three distinct behavioral paths:
\begin{itemize}[itemsep=2pt,topsep=0pt,parsep=0pt]
    \item Path 1 (Rigorous Inference): The model correctly locates the visual anchor, extracts the correct value, and computes the correct answer.
    \item Path 2 (Visual Hallucination): The model hallucinates an incorrect grid coordinate but proceeds to calculate a wrong answer based on the faulty extraction.
    \item Path 3 (Shortcut Guessing): The model bypasses grid localization entirely or extracts incorrect cells, yet exploits language biases to guess the correct final answer.
\end{itemize}
Traditional outcome-supervised RLVR rewards Path 1 and Path 3 equally. Our framework explicitly penalizes Path 3 by isolating the verification of visual anchors from the final answer\cite{cui2025process}.

\paragraph{Process Evaluation via Critic VLM}

We evaluate the fidelity of the generated V-CoT using a separate critic VLM.
A fundamental challenge in training a reliable process verifier is acquiring high-quality negative examples of reasoning trajectories.
To construct the critic's training distribution, we synthesize erroneous V-CoTs by perturbing high-quality reasoning chains.
Specifically, we use an auxiliary 8B-parameter language model (Qwen-3 8B) to systematically corrupt the logical coordinates and intermediate values within human-annotated or verified trajectories.
The severity of the perturbation determines the ground-truth process score.
We then train a 32B-parameter vision-language model (Qwen-3-VL 32B) to function as the critic.
By exposing the critic to both pristine logic and heuristically generated errors, it learns to holistically evaluate the structural integrity of a V-CoT, differentiating between flawless reasoning (Path 1) and varying degrees of structural hallucination (Path 2 and Path 3).

During RL training, for each generated trajectory $y$, the critic VLM produces a dense \emph{process score}, $r_{\text{proc}} \in [0, 1]$.
This score dynamically gates the outcome reward. Let $r_{\text{fmt}} \in [0, 1]$ denote adherence to structural guidelines, and $r_{\text{acc}} \in \{0, 1\}$ indicate final answer correctness. We define the baseline outcome reward as:
\begin{equation}
    R_{\text{base}} = r_{\text{fmt}} + r_{\text{acc}}.
\end{equation}
The final composite reward, $R(y)$, incorporates the critic's feedback via a piecewise gating mechanism, governed by hyperparameters $\alpha$, $\beta$, $\tau_{\text{high}}$, and $\tau_{\text{low}}$:
\begin{equation}
    R(y) =
    \begin{cases}
        R_{\text{base}} + \alpha, & \text{if } R_{\text{base}} > 1 \text{ and } r_{\text{proc}} > \tau_{\text{high}} \\
        r_{\text{fmt}} + \beta, & \text{if } R_{\text{base}} > 1 \text{ and } r_{\text{proc}} < \tau_{\text{low}} \\
        R_{\text{base}} + \alpha \cdot r_{\text{proc}}, & \text{if } R_{\text{base}} > 1 \text{ and } \tau_{\text{low}} \le r_{\text{proc}} \le \tau_{\text{high}} \\
        R_{\text{base}}, & \text{otherwise}.
    \end{cases}
\end{equation}
This formulation yields a Golden-Case Bonus ($\alpha$) for trajectories that achieve both outcome correctness and near-perfect structural reasoning. Conversely, it applies a Conservative Penalty to shortcut guesses (Path 3), reducing the answer credit to $\beta$ to mitigate reward hacking without stifling benign exploration.

\paragraph{Process-Guided Direct Alignment Policy Optimization (PGPO)}

Inspired by the successes of GRPO\cite{shao2024deepseekmath} in text-only reasoning and its subsequent variants, DAPO\cite{yu2025dapo} and LSPO\cite{chen2025lspo}, we optimize the policy VLM using our proposed PGPO algorithm.
PGPO adapts the decoupled clipping bounds of DAPO for stability and the length-aware dynamic sampling of LSPO for efficiency, explicitly tailored for multimodal process supervision.

For a given query $(x, q)$, we sample a group of $G$ reasoning trajectories $\{y_i\}_{i=1}^G$ from the old policy $\pi_{\theta_{\text{old}}}$.
Standard GRPO estimates the advantage $\hat{A}_i$ by normalizing the outcome rewards within the group.
In PGPO, we utilize the composite reward $R(y_i)$ (which includes the critic's process feedback) to compute the advantage:
\begin{equation}
    \hat{A}_i = \frac{R(y_i) - \text{mean}(\{R(y_j)\}_{j=1}^G)}{\text{std}(\{R(y_j)\}_{j=1}^G)}.
\end{equation}

\paragraph{Decoupled Policy Constraints.}
Following DAPO, we decouple the clipping constraints on the importance sampling ratio to stabilize the optimization of long reasoning chains. We define the token-level ratio $\rho_{i,t}(\theta) = \frac{\pi_\theta(y_{i,t} \mid x, q, y_{i,<t})}{\pi_{\theta_{\text{old}}}(y_{i,t} \mid x, q, y_{i,<t})}$. The core DAPO objective maximizes:
{\small
\begin{equation}
    \mathcal{J}_{\text{DAPO}}(\theta) = \frac{1}{G} \sum_{i=1}^G \frac{1}{|y_i|} \sum_{t=1}^{|y_i|} \min \left( \rho_{i,t}(\theta) \hat{A}_i, \text{clip}(\rho_{i,t}(\theta), 1 - \epsilon_{\text{low}}, 1 + \epsilon_{\text{high}}) \hat{A}_i \right),
\end{equation}
}
where asymmetric clipping bounds ($\epsilon_{\text{low}}, \epsilon_{\text{high}}$) prevent premature convergence and mitigate gradient explosion during extended rollouts.

\paragraph{Length-Aware Dynamic Sampling.}
The length of a reasoning trajectory is a strong proxy for problem difficulty and model uncertainty.
In visual reasoning, mid-length trajectories often represent standard pattern matching, whereas extremely short trajectories indicate memorized guesses, and extremely long trajectories reflect complex multi-hop searches or self-correction loops.
Adopting the core insight of LSPO, PGPO introduces a length-aware filtering mechanism at the batch level\cite{wang2025multimodal}.
For the sampled group $\{y_i\}_{i=1}^G$, we compute the sequence lengths $\{|y_i|\}_{i=1}^G$.
Rather than computing gradients over the entire group, we selectively retain a subset $\mathcal{G}_{\text{active}} \subset \{1, \dots, G\}$ comprising trajectories whose lengths fall within the empirically validated thresholds from LSPO (specifically, retaining the bottom 30\% and those between the 60\% and 90\% percentiles).
{\small
\begin{equation}
    \mathcal{J}_{\text{PGPO}}(\theta) = \frac{1}{|\mathcal{G}_{\text{active}}|} 
    \sum_{i \in \mathcal{G}_{\text{active}}} 
    \frac{1}{|y_i|} 
    \sum_{t=1}^{|y_i|} 
    \min \left( 
        \rho_{i,t}(\theta) \hat{A}_i, 
        \text{clip}(\rho_{i,t}(\theta), 1 - \epsilon_{\text{low}}, 1 + \epsilon_{\text{high}}) \hat{A}_i 
    \right).
\end{equation}
}
This dynamic sampling dramatically improves RL efficiency by filtering out low-information, mid-length trajectories. Furthermore, by focusing optimization updates on the longest (most exploratory) and shortest (most decisive) chains, PGPO effectively widens the policy's search distribution, enhancing out-of-domain generalization while avoiding the computational overhead of optimizing redundant reasoning patterns\cite{wang2025visualprm}.

By unifying rigorous visual anchoring, critic-gated process rewards, decoupled clipping, and length-aware sampling, PGPO establishes a mathematically principled and highly efficient framework for scaling multimodal inference.

\section{Experiments}
\label{sec:experiments}

To evaluate the effectiveness of V-tableR1, we construct a comprehensive training and evaluation pipeline across multiple tabular reasoning benchmarks. Our evaluation encompasses two primary tasks: Table Fact Verification (TFV) and Table Question Answering (TQA). 

\subsection{Datasets}
We carefully select a diverse mixture of seven standard tabular reasoning datasets. This selection comprehensively evaluates the model's capability across a wide range of visual table structures, hierarchical headers, and complex reasoning types (such as arithmetic derivation, multi-hop lookups, and semantic entailment). Table~\ref{tab:dataset_stats} provides a detailed statistical overview of the training and evaluation splits, as well as the visual characteristics of the test sets.

\paragraph{Table Fact Verification (TFV).}
\begin{itemize}[itemsep=2pt,topsep=0pt,parsep=0pt]
    \item TabFact\cite{chen2019tabfact}: A large-scale dataset based on Wikipedia tables. The task requires the model to determine whether a given natural language statement is entailed or refuted by the table, testing its capacity for linguistic and symbolic reasoning.
    \item InfoTabs\cite{gupta2020infotabs}: A semi-structured inference dataset built upon Wikipedia infoboxes. It features complex, multi-row logical reasoning where the model must classify hypotheses as entailment, contradiction, or neutral, often requiring implicit knowledge integration.
\end{itemize}

\paragraph{Table Question Answering (TQA).}
\begin{itemize}[itemsep=2pt,topsep=0pt,parsep=0pt]
    \item FinQA\cite{chen2021finqa}: A financial reasoning dataset requiring deep numerical calculation (addition, subtraction, division) over professional earnings reports. It challenges models to extract and compute over numbers heavily embedded in dense tables and accompanying text.
    \item HiTab\cite{cheng2022hitab}: A hierarchical table dataset containing complex statistical reports. It specifically evaluates the model's structural understanding, demanding precise visual anchoring to associate values with their hierarchical row and column headers.
    \item TAT-QA\cite{zhu2021tat}: A hybrid question-answering dataset that involves extracting information from both tables and unstructured text. It assesses a model's ability to seamlessly fuse multimodal and textual evidence for complex financial arithmetic.
    \item TabMWP\cite{lu2022dynamic}: A dataset of tabular math word problems. It requires models to perform multi-step mathematical reasoning explicitly grounded in the contextual tabular data.
    \item WikiTableQuestions (WTQ)\cite{pasupat2015compositional}: A highly compositional dataset over Wikipedia tables. It demands complex queries involving aggregations, superlatives, and multi-hop cell lookups.
\end{itemize}

\begin{table}[t]
  \centering
  \caption{Overview of the tabular reasoning datasets evaluated in this study. The datasets span Table Fact Verification (TFV) and Table Question Answering (TQA). Average image resolution and file size are calculated from the test splits.}
  \label{tab:dataset_stats}
  \resizebox{\linewidth}{!}{
  \begin{tabular}{l ccc cc}
    \toprule
    \multirow{2}{*}{Dataset} & \multicolumn{3}{c}{Number of Samples} & \multicolumn{2}{c}{Image Characteristics (Test)} \\
    \cmidrule(lr){2-4} \cmidrule(lr){5-6}
    & SFT Train & RL Train & Test & Avg. Resolution (W$\times$H) & Avg. Size (KB) \\
    \midrule
    \multicolumn{6}{l}{Table Fact Verification (TFV)} \\
    \midrule
    TabFact & 5,250 & 3,648 & 3,779 & 1360 $\times$ 409 & 103.2 \\
    InfoTabs & - & - & 1,200 & 1425 $\times$ 392 & 60.4 \\
    \midrule
    \multicolumn{6}{l}{Table Question Answering (TQA)} \\
    \midrule
    FinQA & 2,569 & 583 & 1,133 & 1838 $\times$ 308 & 82.4 \\
    HiTab & 3,195 & 1,546 & 1,584 & 1242 $\times$ 458 & 77.6 \\
    TabMWP & 4,412 & 3,791 & 2,686 & 269 $\times$ 192 & 20.7 \\
    WTQ & 5,738 & 5,887 & 2,344 & 1140 $\times$ 520 & 106.9 \\
    TAT-QA & - & - & 736 & 1506 $\times$ 358 & 77.6 \\
    \bottomrule
  \end{tabular}
  }
\end{table}

\begin{table}[t]
  \centering
  \caption{Performance comparison on Table Fact Verification (TFV) and Table Question Answering (TQA) benchmarks. We report the accuracy for each dataset along with the average score for each category. The best performance among open-source models is in \textbf{bold}.}
  \label{tab:main_results}
  \resizebox{\linewidth}{!}{
  \begin{tabular}{l ccc cccccc}
    \toprule
    \multirow{2}{*}{Model} & \multicolumn{3}{c}{TFV} & \multicolumn{6}{c}{TQA} \\
    \cmidrule(lr){2-4} \cmidrule(lr){5-10}
    & TabFact & InfoTabs & Avg. & FinQA & HiTab & TAT-QA & TabMWP & WTQ & Avg. \\
    \midrule
    \multicolumn{10}{l}{Closed-Source VLMs} \\
    \midrule
    GPT-4o & 85.16 & 88.75 & 86.96 & 13.42 & 27.59 & 30.98 & 71.88 & 57.80 & 40.33 \\
    GPT-4.1 & 87.32 & 92.43 & 89.88 & 23.44 & 32.39 & 47.63 & 74.06 & 63.45 & 48.19 \\
    Gemini-2.5-Flash & 88.16 & 91.43 & 89.80 & 40.48 & 55.43 & 70.27 & 88.37 & 78.18 & 66.55 \\
    \midrule
    \multicolumn{10}{l}{Open-Source VLMs} \\
    \midrule
    InternVL-2.5-8B\cite{chen2024expanding} & 72.92 & 74.67 & 73.80 & 19.42 & 32.26 & 41.44 & 72.80 & 57.80 & 44.74 \\
    InternVL-2.5-14B\cite{chen2024expanding} & 75.98 & 77.48 & 76.73 & 24.27 & 33.43 & 46.28 & 74.34 & 61.89 & 48.04 \\
    LLaVA-1.5-7B\cite{liu2024improved} & 23.25 & 32.34 & 27.80 & 1.50 & 3.66 & 8.29 & 5.92 & 12.32 & 6.34 \\
    LLaVA-1.5-13B\cite{liu2024improved} & 28.37 & 37.34 & 32.86 & 1.24 & 5.81 & 10.05 & 6.14 & 15.63 & 7.77 \\
    LLaVA-NeXT-7B\cite{li2024llava} & 34.18 & 41.23 & 37.71 & 2.03 & 4.67 & 12.64 & 19.16 & 8.36 & 9.37 \\
    Qwen2.5-VL-72B\cite{bai2025qwen25vltechnicalreport} & 73.45 & 72.82 & 73.14 & 19.34 & 44.04 & 52.23 & 81.95 & 65.70 & 52.65 \\
    Qwen3-VL-2B\cite{bai2025qwen3} & 70.40 & 74.83 & 72.62 & 10.50 & 29.80 & 32.07 & 68.60 & 46.04 & 37.40 \\
    Qwen3-VL-4B\cite{bai2025qwen3} & 73.28 & 77.33 & 75.31 & 14.74 & 27.75 & 35.19 & 71.92 & 54.72 & 40.86 \\
    Qwen3-VL-8B\cite{bai2025qwen3} & 79.44 & 71.33 & 75.39 & 17.12 & 43.56 & 50.00 & 77.88 & 63.04 & 50.32 \\
    Qwen3-VL-32B\cite{bai2025qwen3} & 82.43 & 78.25 & 80.34 & 26.32 & 46.20 & \textbf{56.23} & 80.37 & 65.49 & 54.92 \\
    QVQ-72B\cite{qvq-72b-preview} & 77.68 & 74.64 & 76.16 & 23.12 & 45.70 & 55.48 & 82.20 & \textbf{68.20} & 54.94 \\
    Table-LLaVA-7B\cite{zheng2024multimodal} & 59.85 & 65.26 & 62.56 & 5.29 & 10.09 & 12.82 & 57.78 & 18.43 & 20.88 \\
    Table-LLaVA-13B\cite{zheng2024multimodal} & 65.00 & 66.91 & 65.96 & 7.71 & 10.85 & 15.67 & 59.77 & 20.41 & 22.88 \\
    \midrule
    \multicolumn{10}{l}{Ours} \\
    \midrule
    Qwen3-VL-2B (SFT) & 79.28 & 85.17 & 82.23 & 15.30 & 36.45 & 43.64 & 74.56 & 47.48 & 43.49 \\
    Qwen3-VL-4B (SFT) & 83.88 & 87.33 & 85.61 & 18.88 & 36.58 & 46.63 & 76.60 & 54.32 & 46.60 \\
    V-tableR1 (2B) & 84.69 & 87.54 & 86.12 & 24.49 & 45.45 & 49.83 & 78.94 & 56.49 & 51.04 \\
    V-tableR1 (4B) & \textbf{87.95} & \textbf{88.94} & \textbf{88.45} & \textbf{28.98} & \textbf{47.24} & 54.23 & \textbf{83.38} & 63.37 & \textbf{55.44} \\
    \bottomrule
  \end{tabular}
  }
\end{table}
\subsection{Main Results}
\label{sec:main_results}

Table~\ref{tab:main_results} presents the performance of V-tableR1 against a comprehensive suite of state-of-the-art vision-language models, ranging from closed-source APIs to large-scale open-weights models. The results demonstrate that explicitly supervising the reasoning process significantly outperforms traditional outcome-supervised approaches, especially in tasks demanding rigorous logical derivation.

\paragraph{Dominance on Complex Question Answering (TQA).}
The TQA benchmarks (FinQA, HiTab, TAT-QA, TabMWP, WTQ) represent the most stringent test of tabular reasoning, requiring models to locate specific cells, perform multi-hop aggregations, and execute arithmetic operations. On these tasks, V-tableR1 establishes a new state-of-the-art among open-source models, despite its relatively compact parameter count.
Notably, V-tableR1 (4B) achieves 28.98\% on FinQA and 83.38\% on TabMWP, outperforming larger models like InternVL-2.5-14B (24.27\% and 74.34\%) and performing highly competitively with the large-scale QVQ-72B (23.12\% and 82.20\%). This confirms our core hypothesis: increasing parameter scale without explicitly grounding the reasoning trajectory is inefficient for complex mathematical and structural tasks. By contrast, V-tableR1 forces the model to output explicit visual anchors before calculation, converting black-box guessing into verifiable arithmetic.

\paragraph{Superiority over Conventional SFT.}
A direct comparison between our intermediate SFT checkpoints and the final RL-tuned models highlights the value of our Process-Guided Direct Alignment Policy Optimization (PGPO).
For instance, moving from Qwen3-VL-2B (SFT) to V-tableR1 (2B) yields absolute gains of +9.19\% on FinQA (15.30\% $\rightarrow$ 24.49\%) and +9.00\% on HiTab (36.45\% $\rightarrow$ 45.45\%). 
Similarly, V-tableR1 (4B) improves over its SFT counterpart by +10.10\% on FinQA and +10.66\% on HiTab (36.58\% $\rightarrow$ 47.24\%).
These substantial jumps prove that while SFT can teach a model the syntax of a reasoning chain, it is the critic-gated process rewards during RL that teach the model to reliably traverse multi-step logic without hallucinating intermediate values.

\paragraph{Competitive Performance on Fact Verification (TFV).}
While the primary design of V-tableR1 targets complex numerical and multi-hop reasoning, it also exhibits exceptional strength in semantic verification. On TabFact and InfoTabs, V-tableR1 (4B) scores 87.95\% and 88.94\% respectively, effectively matching or surpassing the performance of much larger proprietary models like GPT-4.1 (87.32\% / 92.43\%) and Gemini-1.5-Flash (88.16\% / 91.43\%). The improvement over the SFT baseline (e.g., TabFact: 83.88\% $\rightarrow$ 87.95\%) indicates that explicitly verifying cell coordinates also reduces the hallucination of non-existent table facts, leading to more robust semantic entailment.

\paragraph{Efficiency vs. Scale.}
Perhaps the most striking takeaway from Table~\ref{tab:main_results} is the parameter efficiency unlocked by process supervision. Our 4B model consistently outperforms large-scale 72B-parameter models (such as Qwen2.5-VL-72B, which is 18$\times$ larger) across both TFV and TQA tasks. Furthermore, domain-specific SFT models like Table-LLaVA (7B and 13B) perform poorly on TQA tasks (e.g., 5.29\% and 7.71\% on FinQA), highlighting that simply fine-tuning on table data is inadequate. True tabular intelligence requires the structural determinism and step-level penalization that V-tableR1 provides.
\subsection{Ablation Studies}
\label{sec:ablation}

To systematically dissect the source of V-tableR1's performance gains, we conduct ablation studies focusing on the core components of our proposed optimization algorithm. We evaluate the models on both the TFV and TQA benchmarks, using the Qwen3-VL-2B architecture as our base model for these experiments.

\begin{table}[t]
  \centering
  \caption{Ablation study on the components of our proposed optimization framework using the 2B model. We report the average accuracy across TFV and TQA benchmarks.}
  \label{tab:ablation}
  \begin{tabular}{lcc}
    \toprule
    Optimization Method & TFV-Avg. & TQA-Avg. \\
    \midrule
    GRPO & 81.34 & 46.43 \\
    DAPO & 82.48 & 49.04 \\
    PGPO (w/o Process Supervision) & 83.12 & 49.34 \\
    \midrule
    PGPO (Ours) & \textbf{86.12} & \textbf{51.04} \\
    \bottomrule
  \end{tabular}
\end{table}

\paragraph{Impact of Optimization Algorithms.}
Table~\ref{tab:ablation} compares PGPO against standard GRPO, Direct Alignment Policy Optimization (DAPO), and a PGPO variant lacking critic-gated process rewards. DAPO serves as an intermediate baseline using decoupled clipping constraints but lacking length-aware dynamic sampling and process supervision. Upgrading GRPO to DAPO improves the TQA average by 2.61\% (from 46.43\% to 49.04\%), validating that decoupled clipping stabilizes visual domain learning. Adding length-aware dynamic sampling (PGPO w/o Process Supervision) yields a further 0.30\% TQA boost, confirming that filtering low-information, mid-length trajectories enhances optimization efficiency. The most profound leap occurs when activating the critic VLM for step-level process supervision. Full PGPO improves upon GRPO by 4.78\% on TFV (from 81.34\% to 86.12\%) and 4.61\% on TQA (from 46.43\% to 51.04\%). This proves that while clipping and sampling optimizations help, the fundamental visual reasoning bottleneck remains the lack of verifiable, step-level logic grounding.

\begin{wrapfigure}{r}{0.45\linewidth}
  \vspace{-25pt}  
  \centering
  \includegraphics[width=\linewidth]{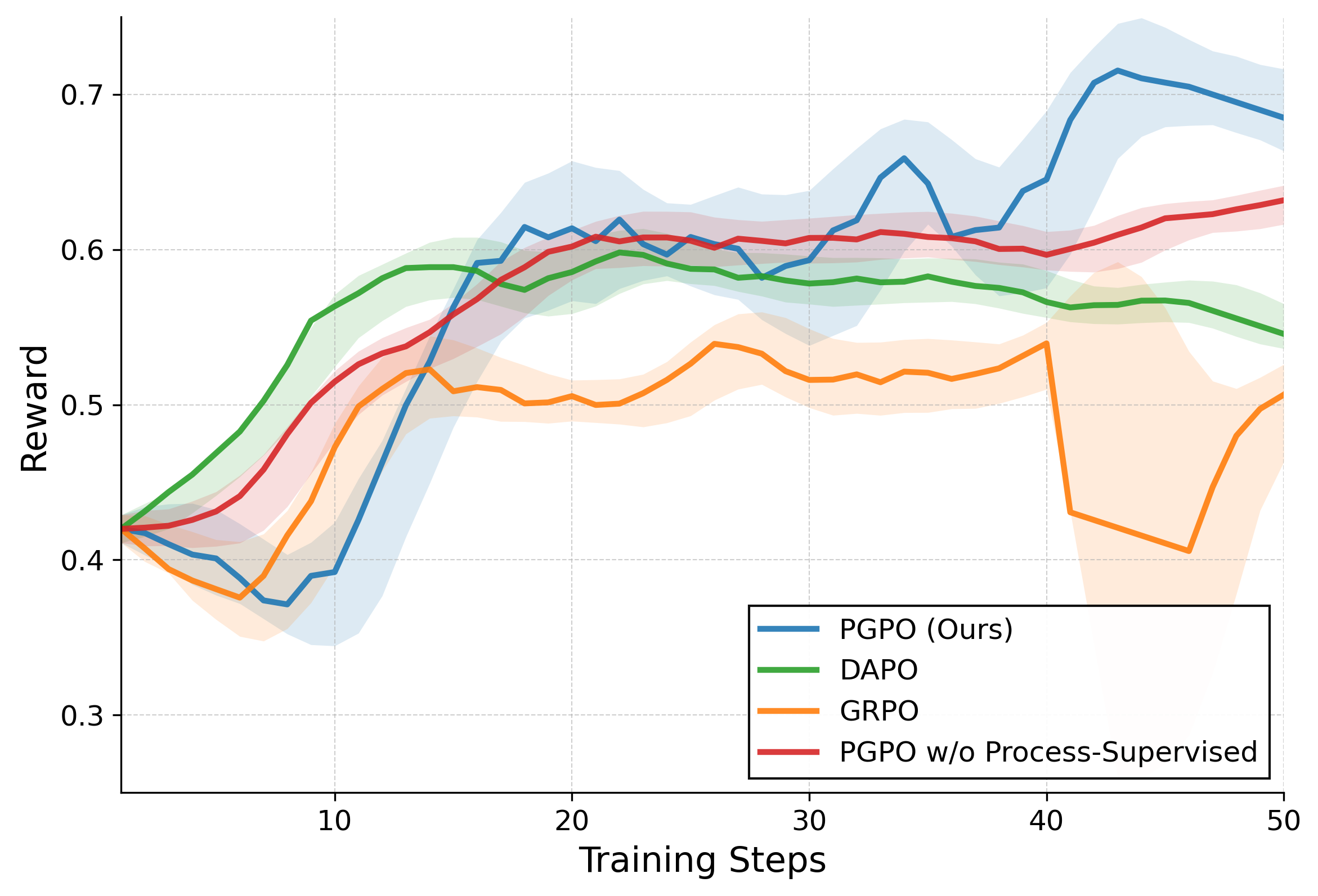}
  \caption{Training reward curves comparing different optimization strategies. Our full PGPO method (blue line) demonstrates superior final convergence and stability compared to GRPO (orange), DAPO (green), and a variant of PGPO lacking process supervision (red).}
  \label{fig:reward_curves}
  \vspace{-25pt}
\end{wrapfigure}

\paragraph{Convergence and Stability Analysis.}
Figure~\ref{fig:reward_curves} visualizes the training reward curves over 50 steps for the evaluated optimization methods. The standard GRPO baseline (orange) exhibits severe instability, highlighted by a catastrophic reward collapse around step 40. This is a classic symptom of reward hacking in visual domains, where the policy exploits a shortcut that suddenly breaks down on slightly out-of-distribution batches.
DAPO (green) mitigates this collapse, showing a smoother curve, but plateaus early, failing to explore deeper logical trajectories. 
Remarkably, our full PGPO method (blue) not only recovers from early exploration dips (steps 5-10) but ultimately climbs to the highest sustained reward level ($>0.7$). The comparison between PGPO and its variant without process supervision (red) is particularly telling: while the outcome-only variant plateaus around 0.6, the full PGPO method breaks through this ceiling. The critic's dense feedback actively steers the policy away from local optima (Shortcut Guessing) and forces it to discover the globally optimal strategy.

\subsection{Reasoning Quality Assessment}
To qualitatively analyze reasoning depth and grounding precision beyond terminal accuracy, we perform an LLM-as-a-judge evaluation using GPT-5. We randomly select 1,000 instances from the combined test sets and task GPT-5 with scoring the generated Chain-of-Thought trajectories of both the Qwen3-VL-2B and V-tableR1-2B. In accordance with established protocols, the judge assigns scores on a scale of 1 to 5 across three critical dimensions: Accuracy, representing the correctness of intermediate factual and numerical steps; Logic, denoting structural soundness and the absence of contradictions; and Visual Grounding, indicating the precision of cell localization and coordinate anchoring. 

As indicated in Table~\ref{tab:gpt_judge}, V-tableR1 improves Visual Grounding from 2.3 to 3.8. This 65\% relative increase validates that process-supervised reinforcement learning with critic feedback directs the model to anchor internal reasoning in correct visual regions. While the baseline has strong logical capabilities at 4.0, its Visual Grounding score of 2.3 highlights a failure to ground logic in pixel space. By penalizing hallucinations and shortcut guessing, V-tableR1 bridges the gap between abstract inference and visual evidence, leading to more reliable reasoning as intermediate accuracy improves from 2.8 to 3.4.
\begin{table}[ht]
\centering
\caption{Reasoning quality assessment by GPT-5 (LLM-as-a-judge) on 1,000 randomly sampled test instances. Scores are on a 1--5 scale.}
\label{tab:gpt_judge}
\begin{tabular}{lccc}
\toprule
\textbf{Model} & \textbf{Accuracy} & \textbf{Logic} & \textbf{Visual Grounding} \\
\midrule
Qwen3-VL-2B (Baseline) & 2.8 & 4.0 & 2.3 \\
V-tableR1-2B & \textbf{3.4} & \textbf{4.2} & \textbf{3.8} \\
\bottomrule
\end{tabular}
\end{table}

\subsection{Qualitative Analysis}
\label{sec:qualitative}

Illustrating process supervision impact, Figure~\ref{fig:reward_curves} contrasts reasoning trajectories of the outcome-supervised pre-trained baseline and our PGPO-trained V-tableR1 on a multi-hop structural grounding query.

\begin{figure}[t]
\centering
\includegraphics[width=\linewidth]{P-3.png}
\caption{Qualitative trajectory comparison pre- and post-PGPO. When querying total rushing yards, the baseline (top right) suffers Visual Hallucination, mistakenly extracting the receiving section value. Conversely, V-tableR1 (bottom right) ensures Rigorous Inference and correct answering via an explicit visual anchor (<cell: Row 14, Col 4>).}
\label{fig:case_study}
\end{figure}

\textbf{Baseline Failure.}
Querying Craig eleven-season total rushing yards, the standard Qwen3-VL-2B baseline correctly parses semantic intent but executes ungrounded reasoning (Figure~\ref{fig:reward_curves}, top right). It outputs 4911.0 after erroneously referencing 4911 receiving yards. Illustrating Visual Hallucination (Path 2), attention drifts to the receiving super-header, extracting the incorrect yds total. Lacking fine-grained process supervision, the baseline escapes penalization for failing to anchor attention to the correct visual bounding box.

\textbf{V-tableR1 Success.}
Post-PGPO, dense critic rewards compel explicit visual anchor generation and verification before value extraction (Figure~\ref{fig:case_study}, bottom right). The model correctly identifies the rushing section, the yds column (Column 4), and the bottom row totals (Row 14). Extracting the intersection value <cell: Row 14, Col 4>, it outputs the correct answer 8189.0. Exemplifying Rigorous Inference (Path 1), PGPO forces explicit grid coordinate generation, inextricably binding logical operations to the correct visual region and eliminating baseline spatial cross-talk.

\section{Conclusion}
\label{sec:conclusion}
In this paper, we identified a critical flaw in current vision-language models: the reliance on outcome-supervised training, which inadvertently encourages visual hallucinations and shortcut guessing during multi-step reasoning. To address this, we introduced V-tableR1, a framework that forces models to explicitly generate and verify visual chain-of-thought anchors. At the core of our approach is Process-Guided Direct Alignment Policy Optimization, a novel RL algorithm that integrates decoupled policy constraints, length-aware dynamic sampling, and most importantly, dense step-level feedback from a critic VLM. Our extensive evaluations across seven complex tabular benchmarks prove that process supervision fundamentally shifts multimodal inference from opaque pattern matching to rigorous logical derivation. The resulting 4B-parameter V-tableR1 achieves state-of-the-art performance, outperforming massive 72B-parameter models and establishing a new paradigm for structured visual reasoning.

%
%
\bibliographystyle{splncs04}
\bibliography{main}

\begin{thebibliography}{10}
\providecommand{\url}[1]{\texttt{#1}}
\providecommand{\urlprefix}{URL }
\providecommand{\doi}[1]{https://doi.org/#1}

\bibitem{antol2015vqa}
Antol, S., Agrawal, A., Lu, J., Mitchell, M., Batra, D., Zitnick, C.L., Parikh, D.: Vqa: Visual question answering. In: Proceedings of the IEEE international conference on computer vision. pp. 2425--2433 (2015)

\bibitem{bai2025qwen3}
Bai, S., Cai, Y., Chen, R., Chen, K., Chen, X., Cheng, Z., Deng, L., Ding, W., Gao, C., Ge, C., et~al.: Qwen3-vl technical report. arXiv preprint arXiv:2511.21631  (2025)

\bibitem{bai2025qwen25vltechnicalreport}
Bai, S., Chen, K., Liu, X., Wang, J., Ge, W., Song, S., Dang, K., Wang, P., Wang, S., Tang, J., Zhong, H., Zhu, Y., Yang, M., Li, Z., Wan, J., Wang, P., Ding, W., Fu, Z., Xu, Y., Ye, J., Zhang, X., Xie, T., Cheng, Z., Zhang, H., Yang, Z., Xu, H., Lin, J.: Qwen2.5-vl technical report (2025), \url{https://arxiv.org/abs/2502.13923}

\bibitem{chen2025lspo}
Chen, W., Koenig, S., Dilkina, B.: Lspo: Length-aware dynamic sampling for policy optimization in llm reasoning. arXiv preprint arXiv:2510.01459  (2025)

\bibitem{chen2019tabfact}
Chen, W., Wang, H., Chen, J., Zhang, Y., Wang, H., Li, S., Zhou, X., Wang, W.Y.: Tabfact: A large-scale dataset for table-based fact verification. arXiv preprint arXiv:1909.02164  (2019)

\bibitem{chen2024expanding}
Chen, Z., Wang, W., Cao, Y., Liu, Y., Gao, Z., Cui, E., Zhu, J., Ye, S., Tian, H., Liu, Z., et~al.: Expanding performance boundaries of open-source multimodal models with model, data, and test-time scaling. arXiv preprint arXiv:2412.05271  (2024)

\bibitem{chen2024visual}
Chen, Z., Zhou, Q., Shen, Y., Hong, Y., Sun, Z., Gutfreund, D., Gan, C.: Visual chain-of-thought prompting for knowledge-based visual reasoning. In: Proceedings of the AAAI Conference on Artificial Intelligence. vol.~38, pp. 1254--1262 (2024)

\bibitem{chen2021finqa}
Chen, Z., Chen, W., Smiley, C., Shah, S., Borova, I., Langdon, D., Moussa, R., Beane, M., Huang, T.H., Routledge, B.R., et~al.: Finqa: A dataset of numerical reasoning over financial data. In: Proceedings of the 2021 Conference on Empirical Methods in Natural Language Processing. pp. 3697--3711 (2021)

\bibitem{cheng2022hitab}
Cheng, Z., Dong, H., Wang, Z., Jia, R., Guo, J., Gao, Y., Han, S., Lou, J.G., Zhang, D.: Hitab: A hierarchical table dataset for question answering and natural language generation. In: Proceedings of the 60th Annual Meeting of the Association for Computational Linguistics (Volume 1: Long Papers). pp. 1094--1110 (2022)

\bibitem{cui2025process}
Cui, G., Yuan, L., Wang, Z., Wang, H., Zhang, Y., Chen, J., Li, W., He, B., Fan, Y., Yu, T., et~al.: Process reinforcement through implicit rewards. arXiv preprint arXiv:2502.01456  (2025)

\bibitem{favero2024multi}
Favero, A., Zancato, L., Trager, M., Choudhary, S., Perera, P., Achille, A., Swaminathan, A., Soatto, S.: Multi-modal hallucination control by visual information grounding. In: Proceedings of the IEEE/CVF Conference on Computer Vision and Pattern Recognition. pp. 14303--14312 (2024)

\bibitem{geirhos2020shortcut}
Geirhos, R., Jacobsen, J.H., Michaelis, C., Zemel, R., Brendel, W., Bethge, M., Wichmann, F.A.: Shortcut learning in deep neural networks. Nature Machine Intelligence  \textbf{2}(11),  665--673 (2020)

\bibitem{gupta2020infotabs}
Gupta, V., Mehta, M., Nokhiz, P., Srikumar, V.: Infotabs: Inference on tables as semi-structured data. In: Proceedings of the 58th Annual Meeting of the Association for Computational Linguistics. pp. 2309--2324 (2020)

\bibitem{ji2024tree}
Ji, D., Zhu, L., Gao, S., Xu, P., Lu, H., Ye, J., Zhao, F.: Tree-of-table: Unleashing the power of llms for enhanced large-scale table understanding. arXiv preprint arXiv:2411.08516  (2024)

\bibitem{jiao2024learning}
Jiao, F., Qin, C., Liu, Z., Chen, N., Joty, S.: Learning planning-based reasoning by trajectories collection and process reward synthesizing. In: Proceedings of the 2024 Conference on Empirical Methods in Natural Language Processing. pp. 334--350 (2024)

\bibitem{laurenccon2024matters}
Lauren{\c{c}}on, H., Tronchon, L., Cord, M., Sanh, V.: What matters when building vision-language models? Advances in Neural Information Processing Systems  \textbf{37},  87874--87907 (2024)

\bibitem{li2024llava}
Li, F., Zhang, R., Zhang, H., Zhang, Y., Li, B., Li, W., Ma, Z., Li, C.: Llava-next-interleave: Tackling multi-image, video, and 3d in large multimodal models. arXiv preprint arXiv:2407.07895  (2024)

\bibitem{liu2024improved}
Liu, H., Li, C., Li, Y., Lee, Y.J.: Improved baselines with visual instruction tuning. In: Proceedings of the IEEE/CVF conference on computer vision and pattern recognition. pp. 26296--26306 (2024)

\bibitem{lu2022learn}
Lu, P., Mishra, S., Xia, T., Qiu, L., Chang, K.W., Zhu, S.C., Tafjord, O., Clark, P., Kalyan, A.: Learn to explain: Multimodal reasoning via thought chains for science question answering. Advances in neural information processing systems  \textbf{35},  2507--2521 (2022)

\bibitem{lu2022dynamic}
Lu, P., Qiu, L., Chang, K.W., Wu, Y.N., Zhu, S.C., Rajpurohit, T., Clark, P., Kalyan, A.: Dynamic prompt learning via policy gradient for semi-structured mathematical reasoning. arXiv preprint arXiv:2209.14610  (2022)

\bibitem{mroueh2025reinforcement}
Mroueh, Y.: Reinforcement learning with verifiable rewards: Grpo's effective loss, dynamics, and success amplification. arXiv preprint arXiv:2503.06639  (2025)

\bibitem{pasupat2015compositional}
Pasupat, P., Liang, P.: Compositional semantic parsing on semi-structured tables. In: Proceedings of the 53rd Annual Meeting of the Association for Computational Linguistics and the 7th International Joint Conference on Natural Language Processing (Volume 1: Long Papers). pp. 1470--1480 (2015)

\bibitem{rawte2025defining}
Rawte, V., Mishra, A., Sheth, A., Das, A.: Defining and quantifying visual hallucinations in vision-language models. In: Proceedings of the 5th Workshop on Trustworthy NLP (TrustNLP 2025). pp. 501--510 (2025)

\bibitem{robinson2021can}
Robinson, J., Sun, L., Yu, K., Batmanghelich, K., Jegelka, S., Sra, S.: Can contrastive learning avoid shortcut solutions? Advances in neural information processing systems  \textbf{34},  4974--4986 (2021)

\bibitem{shao2024deepseekmath}
Shao, Z., Wang, P., Zhu, Q., Xu, R., Song, J., Bi, X., Zhang, H., Zhang, M., Li, Y., Wu, Y., et~al.: Deepseekmath: Pushing the limits of mathematical reasoning in open language models. arXiv preprint arXiv:2402.03300  (2024)

\bibitem{su2025crossing}
Su, Y., Yu, D., Song, L., Li, J., Mi, H., Tu, Z., Zhang, M., Yu, D.: Crossing the reward bridge: Expanding rl with verifiable rewards across diverse domains. arXiv preprint arXiv:2503.23829  (2025)

\bibitem{sun2020tableqa}
Sun, N., Yang, X., Liu, Y.: Tableqa: a large-scale chinese text-to-sql dataset for table-aware sql generation. arXiv preprint arXiv:2006.06434  (2020)

\bibitem{qvq-72b-preview}
Team, Q.: Qvq: To see the world with wisdom (December 2024), \url{https://qwenlm.github.io/blog/qvq-72b-preview/}

\bibitem{titiya2025mmtbench}
Titiya, P.Y., Trivedi, J., Baral, C., Gupta, V.: Mmtbench: A unified benchmark for complex multimodal table reasoning. arXiv preprint arXiv:2505.21771  (2025)

\bibitem{vojnovic2025alignment}
Vojnovic, M., Yun, S.Y.: What is the alignment objective of grpo? arXiv preprint arXiv:2502.18548  (2025)

\bibitem{wang2024rethinking}
Wang, Q., Wang, Z., Su, Y., Tong, H., Song, Y.: Rethinking the bounds of llm reasoning: Are multi-agent discussions the key? In: Proceedings of the 62nd Annual Meeting of the Association for Computational Linguistics (Volume 1: Long Papers). pp. 6106--6131 (2024)

\bibitem{wang2025visualprm}
Wang, W., Gao, Z., Chen, L., Chen, Z., Zhu, J., Zhao, X., Liu, Y., Cao, Y., Ye, S., Zhu, X., et~al.: Visualprm: An effective process reward model for multimodal reasoning. arXiv preprint arXiv:2503.10291  (2025)

\bibitem{wang2025multimodal}
Wang, Y., Wu, S., Zhang, Y., Yan, S., Liu, Z., Luo, J., Fei, H.: Multimodal chain-of-thought reasoning: A comprehensive survey. arXiv preprint arXiv:2503.12605  (2025)

\bibitem{wang2024chain}
Wang, Z., Zhang, H., Li, C.L., Eisenschlos, J.M., Perot, V., Wang, Z., Miculicich, L., Fujii, Y., Shang, J., Lee, C.Y., et~al.: Chain-of-table: Evolving tables in the reasoning chain for table understanding. arXiv preprint arXiv:2401.04398  (2024)

\bibitem{wu2025table}
Wu, Z., Yang, J., Liu, J., Wu, X., Pan, C., Zhang, J., Zhao, Y., Song, S., Li, Y., Li, Z.: Table-r1: Region-based reinforcement learning for table understanding. arXiv preprint arXiv:2505.12415  (2025)

\bibitem{yang2022tableformer}
Yang, J., Gupta, A., Upadhyay, S., He, L., Goel, R., Paul, S.: Tableformer: Robust transformer modeling for table-text encoding. In: Proceedings of the 60th Annual Meeting of the Association for Computational Linguistics (Volume 1: Long Papers). pp. 528--537 (2022)

\bibitem{yu2025dapo}
Yu, Q., Zhang, Z., Zhu, R., Yuan, Y., Zuo, X., Yue, Y., Dai, W., Fan, T., Liu, G., Liu, L., et~al.: Dapo: An open-source llm reinforcement learning system at scale. arXiv preprint arXiv:2503.14476  (2025)

\bibitem{yue2025vapo}
Yue, Y., Yuan, Y., Yu, Q., Zuo, X., Zhu, R., Xu, W., Chen, J., Wang, C., Fan, T., Du, Z., et~al.: Vapo: Efficient and reliable reinforcement learning for advanced reasoning tasks. arXiv preprint arXiv:2504.05118  (2025)

\bibitem{zhang2024vision}
Zhang, J., Huang, J., Jin, S., Lu, S.: Vision-language models for vision tasks: A survey. IEEE transactions on pattern analysis and machine intelligence  \textbf{46}(8),  5625--5644 (2024)

\bibitem{zhang2025survey}
Zhang, X., Wang, D., Dou, L., Zhu, Q., Che, W.: A survey of table reasoning with large language models. Frontiers of Computer Science  \textbf{19}(9),  199348 (2025)

\bibitem{zhang2025lessons}
Zhang, Z., Zheng, C., Wu, Y., Zhang, B., Lin, R., Yu, B., Liu, D., Zhou, J., Lin, J.: The lessons of developing process reward models in mathematical reasoning. In: Findings of the Association for Computational Linguistics: ACL 2025. pp. 10495--10516 (2025)

\bibitem{zhang2023multimodal}
Zhang, Z., Zhang, A., Li, M., Zhao, H., Karypis, G., Smola, A.: Multimodal chain-of-thought reasoning in language models. arXiv preprint arXiv:2302.00923  (2023)

\bibitem{zhao2025cot}
Zhao, Q., Lu, Y., Kim, M.J., Fu, Z., Zhang, Z., Wu, Y., Li, Z., Ma, Q., Han, S., Finn, C., et~al.: Cot-vla: Visual chain-of-thought reasoning for vision-language-action models. In: Proceedings of the Computer Vision and Pattern Recognition Conference. pp. 1702--1713 (2025)

\bibitem{zhao2024tabpedia}
Zhao, W., Feng, H., Liu, Q., Tang, J., Wu, B., Liao, L., Wei, S., Ye, Y., Liu, H., Zhou, W., et~al.: Tabpedia: Towards comprehensive visual table understanding with concept synergy. Advances in Neural Information Processing Systems  \textbf{37},  7185--7212 (2024)

\bibitem{zheng2024multimodal}
Zheng, M., Feng, X., Si, Q., She, Q., Lin, Z., Jiang, W., Wang, W.: Multimodal table understanding. In: Proceedings of the 62nd Annual Meeting of the Association for Computational Linguistics (Volume 1: Long Papers). pp. 9102--9124 (2024)

\bibitem{zhou2022learning}
Zhou, K., Yang, J., Loy, C.C., Liu, Z.: Learning to prompt for vision-language models. International journal of computer vision  \textbf{130}(9),  2337--2348 (2022)

\bibitem{zhu2023minigpt}
Zhu, D., Chen, J., Shen, X., Li, X., Elhoseiny, M.: Minigpt-4: Enhancing vision-language understanding with advanced large language models. arXiv preprint arXiv:2304.10592  (2023)

\bibitem{zhu2021tat}
Zhu, F., Lei, W., Huang, Y., Wang, C., Zhang, S., Lv, J., Feng, F., Chua, T.S.: Tat-qa: A question answering benchmark on a hybrid of tabular and textual content in finance. In: Proceedings of the 59th annual meeting of the Association for Computational Linguistics and the 11th international joint conference on natural language processing (volume 1: long papers). pp. 3277--3287 (2021)

\end{thebibliography}
\end{document}